\documentclass[10pt]{article}

\usepackage[letterpaper, margin=0.9in]{geometry}

\usepackage{times}          % Times New Roman
\usepackage[T1]{fontenc}

\usepackage{xcolor}
\usepackage{natbib}
\usepackage{graphicx} % Required for inserting images
\usepackage{amsmath}
\usepackage{amsfonts}
\usepackage{amssymb}
\usepackage{mathtools}    % Extensions to amsmath
\usepackage{bm}           % Bold math symbols
\usepackage{bbm}          % Indicator functions if needed
\usepackage{graphicx}     % Figures
\usepackage{booktabs}     % Professional tables
\usepackage{enumitem}     % Control over lists (used in Scenarios section)
\usepackage{hyperref}     % Cross-referencing
\usepackage{subcaption}

 \usepackage{setspace}
\usepackage{ragged2e}
\justifying

\usepackage{titlesec}

\titleformat{\section}
  {\normalfont\bfseries\fontsize{12}{14}\selectfont}
  {\thesection.}{0.5em}{}

\titleformat{\subsection}
  {\normalfont\bfseries\fontsize{10}{12}\selectfont}
  {\thesubsection.}{0.5em}{}

\titlespacing*{\section}{0pt}{12pt}{6pt}
\titlespacing*{\subsection}{0pt}{6pt}{6pt}

\title{Accelerating Reinforcement Learning Training Using Simulation Surrogate Models}
\date{}

\begin{document}
\author{
Mohammadmahdi Ghasemloo\textsuperscript{1},
David J. Eckman\textsuperscript{1},
Yaxian Li\textsuperscript{2}\\[0.3cm]
\textsuperscript{1}Department of Industrial and Systems Engineering, Texas A\&M University, College Station, TX, USA\\
\textsuperscript{2}Intuit AI, Mountain View, CA, USA
}

\maketitle
\begin{abstract}

{\small High-fidelity simulation models are widely used to analyze complex stochastic systems, but their high computational cost motivates the development of cheaper surrogate models that approximate the simulation model's input-output relationship.
In parallel, reinforcement learning (RL) has emerged as a powerful framework for making online decisions in stochastic environments, with increasing attention being given to the use of simulation models as training environments for RL models.
We investigate a class of surrogate models suitable for accelerating RL training in settings where the reward structure, model parameters, or system dynamics change over time and explore their interactions with simulation models and RL models.
Through numerical experiments on a stochastic service system modeled via discrete-event simulation, we demonstrate that leveraging surrogate models can substantially accelerate RL training and re-training. }
\end{abstract}

\section*{Keywords}
Reinforcement learning, surrogate modeling, discrete-event simulation

\section{Introduction}

Discrete-event simulation (DES) has long been a fundamental tool for analyzing complex stochastic systems that are analytically intractable.
Simulation models are widely used for performance evaluation, what-if analysis, and simulation optimization in manufacturing, service systems, logistics, and healthcare \cite{banks2005discrete, law2014simulation}.
Despite their modeling power, simulation models are often computationally expensive and may require substantial computational effort to execute a single replication, particularly when system dynamics are complex or long-run performance measures are of interest.
This computational burden becomes a critical limitation when simulation models are used within iterative decision making frameworks that require a large number of replications \citep{barton2020tutorial,do2022metamodel}. A common approach to alleviate this issue is to construct a surrogate model (metamodel) that approximates the mapping of simulation inputs to outputs for use in lieu of expensive simulation runs during analysis or optimization \cite{kleijnen2015design}. The original simulation is run at a chosen set of input configurations, and a model of the response surface is fitted to the observed outputs using regression, e.g., polynomial models, stochastic kriging, or neural networks \cite{AlHindi2004, Kilmer1994, Ankenman2010}. The surrogate model is subsequently used to predict outputs for new, unsimulated inputs. Advances in handling integer-valued inputs and uncertainty-aware neural network surrogates have further broadened the applicability of surrogate models \cite{cen2022enhanced}.

In parallel, reinforcement learning (RL) has emerged as a powerful framework for sequential decision making in stochastic environments.
In RL, an agent interacts with an environment over time, observes system states, selects actions, and receives rewards with the objective of learning a policy that maximizes the expected cumulative reward.
Recent advances in function approximation and deep learning have significantly expanded the applicability of RL to large-scale and high-dimensional problems \cite{powell2022reinforcement}.

A growing literature has integrated RL with DES by using DES models as training environments for RL models.
This approach enables agents to learn policies without interacting with real systems.
~\cite{zhang1995reinforcement} apply RL to the Job Shop Scheduling Problem (JSSP) where scheduling is framed as a sequential decision process in which actions correspond to local repair operations.
~\cite{kintsakis2019reinforcement} propose an RL scheduler for workflow management systems operating over distributed computing resources where a policy-gradient-based RL model is trained to assign tasks to heterogeneous resources.
~\cite{bhattathiri2023simulation} introduce a Deep Q-Network (DQN)-based dispatching framework for autonomous mobile robots in warehouse logistics, trained with a DES developed using SimPy that improves on-time deliveries and demonstrates policy generalization across layouts.
Similarly, ~\cite{soykan2024optimizing} integrate Double DQN with DES to solve the JSSP, showing improvements in makespan and machine utilization over priority dispatching rules.
Applications of RL further extend to traditional agent-based simulation approaches that rely on hard-coded behavioral rules by enabling adaptive, data-driven agent behaviors \cite{zhang2021synergistic}.
In agent-based simulation, RL algorithms are commonly used to model agent behavior \cite{lee2017agent} and can further enhance the modeling power through adaptive decision making \cite{li2019agent}.

% RL training typically requires a large number of environment interactions, making direct application to expensive simulation models computationally prohibitive.
The RL literature has proposed several methods to accelerate training.
Model-based RL approaches learn approximate environment dynamics known as world models to enable faster planning and RL training \cite{hafner2019learning}.
Other techniques improve sample efficiency and scalability, including prioritized experience replay \cite{schaul2015prioritized} and distributed RL architectures such as IMPALA \cite{espeholt2018impala}.
Although world models have been extensively studied, less attention has been given to the interaction between simulation surrogate models and RL models in settings where a high-fidelity simulation model is already available for decision making. 
% \textcolor{blue}{May need to clarify the terminology earlier in the introduction. Some would view the simulation as a "surrogate" for the real world.}

This paper makes the following contributions: First, we articulate a problem setting in which a high-fidelity simulation model serves as the ``world model,'' and a surrogate model is constructed to approximate the simulation. Second, we discuss the structural properties a surrogate must satisfy in order to interact effectively with an RL agent. Finally, we demonstrate how such surrogates can be leveraged to accelerate RL training in systems modeled using DES.

\section{Problem Formulation}
\label{sec:math}
Surrogate models for simulation can be constructed in several ways. First, a surrogate may be designed to output a summary statistic of a KPI, such as its expected value.
Second, a surrogate may approximate the full distribution of a KPI.
Third, a surrogate may generate stochastic outputs at intermediate decision epochs, producing random samples that emulate the evolution of the simulation over time.
We focus exclusively on the third class of surrogate models that can interact with an RL model by serving as a training environment. 
% We first introduce mathematical notation to formalize our discussion.

\subsection{Model Representation}

The KPI of a stochastic simulation model can generally be written as $Y(\mathbf{x}) = \mu(\mathbf{x}) + \epsilon(\mathbf{x})$, where $\mathbf{x}$ is a vector of contextual of controllable variables, $\mu$ is referred to as the response surface, and $\epsilon$ is mean-zero noise. Traditionally, surrogate models have attempted to model $\mu$. More modern surrogate models feature generative models that produce random realizations of $Y$ using approaches such as Conditional Variational Autoencoders (CVAEs). When RL agents are embedded in the simulation model and make decisions sequentially, one may be interested in leveraging surrogate models in which outputs are generated sequentially over epochs (referred to as episodes in the RL literature) for computational purposes.

Let $\mathbf{s}_j \in \mathcal{S}$ denote the full system state at the beginning of epoch $j$ for $j=1, \ldots, T$. Assume $\mathcal{M}$ is the set of the state variables (e.g., the queue lengths, utilization of experts), and hence  $\mathcal{S} \subseteq \mathbb{R}^{|\mathcal{M}|}$.
In the RL model embedded in the real system, decisions are made dynamically with its set of state variables and the corresponding state space denoted by $\mathcal{M}' \subseteq \mathcal{M}$ and $\mathcal{S}' \subseteq \mathbb{R}^{|\mathcal{M}'|}$. At the beginning of each epoch $j$, the policy selects an action $a_j \in \mathcal{A}$
and interacts with the environment (i.e., the simulation model) to generate an output $(\mathbf{y}_j, \mathbf{s}_{j+1})$ from the distribution of 
\begin{equation}
\label{eq_main}
Y_j, S_{j+1} \mid (S_{j}=\mathbf{s}_{j},~ A_{j}=\mathbf{a}_{j})
\end{equation}
where a scalar reward is then computed as a function of the simulation output $y_j \in \mathcal{Y}$.

The surrogate model is constructed to approximate the simulation output. The set of state variables that the surrogate model uses is $\mathcal{M''} \subseteq \mathcal{M}$ and the space of states it utilizes is denoted as $\mathcal{S''} \subseteq \mathbb{R}^{|\mathcal{M''}|}$.
At each epoch $j$, the surrogate generates a random sample $(\tilde{\mathbf{s}}_{j+1}, \tilde{\mathbf{y}}_j)$
which is approximately distributed as (\ref{eq_main}). The input space of the surrogate model is $\mathcal{S''} \times \mathcal{A}$ and the output space is $\mathcal{S''} \times \mathcal{Y} $,
with $\mathcal{M}' \subseteq \mathcal{M''}$. This restriction on $\mathcal{M''}$ arises from the requirement that, in order to interact with the RL model, the surrogate must be capable of generating at least the set of state features used as inputs by the RL agent. The surrogate model should therefore be a generative model capable of producing stochastic outputs. Approaches such as combining CVAEs with Long Short-Term Memory (LSTM) cells have been proposed to handle this setting \cite{cen2022enhanced}. 
%Having identified and discussed the class of surrogate models that can be used to interact with the RL model, 
We next explore applications in which surrogate models that can interact with an RL model can be beneficial.

\subsection{Applications}

The discussed class of surrogate models can be constructed in two ways.

\begin{figure}[htbp]
    \centering
    \begin{subfigure}[t]{0.41\linewidth}
        \centering
        \includegraphics[width=\linewidth]{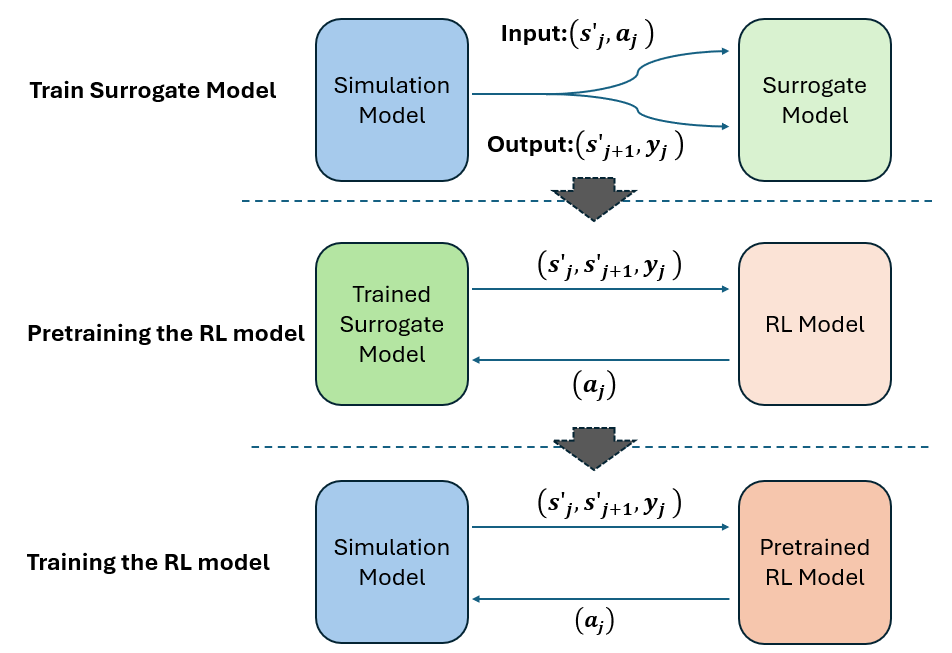}
        \caption{\textbf{Surrogate model trained separately from RL model.}}
        \label{fig:sep}
    \end{subfigure}
    \hfill
    \begin{subfigure}[t]{0.57\linewidth}
        \centering
        \includegraphics[width=\linewidth]{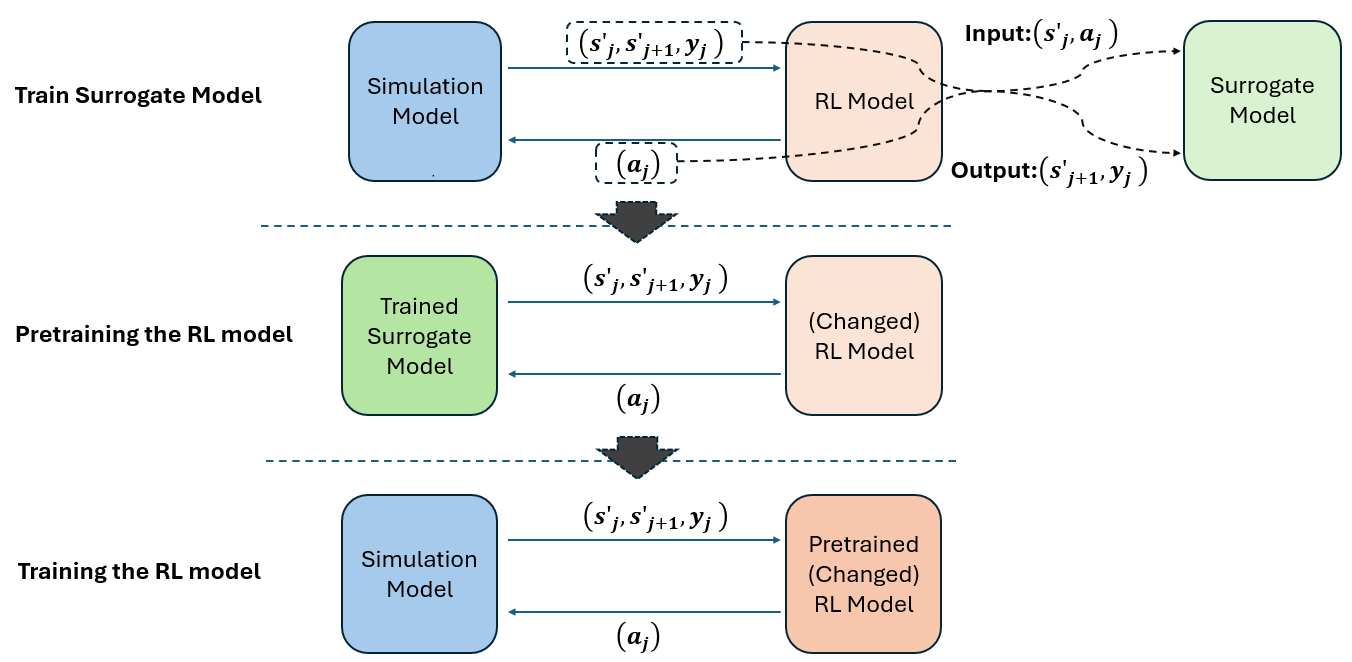}
        \caption{\textbf{Surrogate model trained on the RL trajectories.}}
        \label{fig:der}
    \end{subfigure}

    \caption{Applications in which a simulation surrogate model interacts with the
RL model.}
    \label{fig:sepder}
\end{figure}

\begin{enumerate}
\item \textbf{Surrogate model trained separately.}  
In some cases, the surrogate models are constructed separately from their eventual use in RL training. 
Such a surrogate can later be incorporated into the RL training pipeline to reduce computational cost, for example, by training or warm-starting the RL model. In other words, the surrogate model can be used to initialize the parameters of the RL model, which is then either deployed directly or further fine-tuned through interaction with the high-fidelity simulation model; see Figure~\ref{fig:sep}.

\item \textbf{RL-first training with surrogate learned from RL trajectories.}  
An RL model is first trained directly using the simulation model.
During this process, state-action-output trajectories ($\mathbf{s}_{j}, \mathbf{a}_{j}, \mathbf{y}_j, \mathbf{s}_{j+1}$) generated by the RL agent interacting with the simulation model are collected.
A surrogate model is then trained using this data to approximate the simulation response in regions of the state-action space visited by the RL policy; see Figure~\ref{fig:der}.
While this surrogate is not intended for direct decision making, it can be used to accelerate retraining of the RL model in applications where the RL formulation changes (e.g., reward modification or policy architecture updates).

\end{enumerate}

Surrogate models can then be integrated into the RL training pipeline to address changes in the RL model or changes in system parameters. Examples include:

\begin{itemize}
\item \textbf{Reward function change.}  
Suppose a surrogate model and an RL policy have been trained, and the reward function is subsequently modified.
Such changes commonly arise in practice; for example, a call-center manager may adjust the reward structure to place greater importance on a specific customer class.
Since retraining an RL model directly using the simulation model could be computationally expensive, the surrogate model could be used to rapidly retrain the RL policy under the modified reward function.

\item \textbf{RL model structure change.}  
Changes in the RL model architecture or training algorithm (e.g., the learning algorithm, number of network layers, or feature representation) may require retraining the RL policy.
In such cases, the surrogate model can be used either as a stand-alone training environment or to initialize the parameters of the modified RL model prior to fine tuning using the simulation model.

\item \textbf{Structural changes in the system.}  
The structure of the simulated system may change over time due to capacity adjustments or logic modifications.
A surrogate model can be constructed to capture a broad range of such structural variations and serve as an intermediate environment for efficiently updating the RL policy.

\item \textbf{Changes in input models.}  
Changes in the stochastic input models (e.g., arrival processes or service-time distributions) may alter system behavior.
If the surrogate model is designed to capture such distributional changes, it can be used to update or retrain the RL policy in response to detected input model shifts.
\end{itemize}
% , while the RL model lacks this level of adaptability,
A surrogate model’s input and output representation determines which applications it can effectively support, with more expressive surrogates enabling a broader range of interactions between the simulation and RL models.

\section{Numerical Experiments}
\label{sec:exp}
We consider a discrete-event simulation of a call-center system with two contact groups and three expert groups.
The numbers of experts in the three expert groups are 1, 2, and 1, respectively.
Customers in the two contact groups arrive according to independent Poisson processes with rates
$\lambda_1 = 7$ and $\lambda_2 = 6$. Each customer is characterized by both a service time and a patience time. A customer will abandon the queue if their waiting time exceeds their patience threshold.
The patience- and service-time distributions for the contact groups are $ \text{Gamma}(5, 0.9)$ and $\text{Gamma}(4, 1)$ for contact group~1 and $\text{Gamma}(2, 5)$ and $\text{Gamma}(4, 1.5)$ for contact group~2.

Each expert is assigned a fixed number of back-office tasks with task durations following a lognormal distribution with mean and variance both set at 1.7.
At each decision epoch (each 30-minute interval), expert $k$ selects an action $a_k \in \{0,1\}$, where $0$ indicates front-office service and $1$ indicates back-office work.
Deep Q-Learning (DQN) is used where RL agents operate on a $7$-dimensional observation space and a 4-dimensional action set.
The network consists of two fully connected hidden layers with $32$ neurons.
Training uses learning rate $\alpha = 10^{-4}$, replay memory size $300$, mini-batch size $5$, exploration rate $\epsilon = 0.05$, and discount factor $\gamma = 0.9$.
At each epoch, the simulation returns four KPI vectors: contact groups' average waiting times and abandonment rates, and expert utilizations and the numbers of remaining back-office tasks.
Let $\mathbf{W}_j = (W_{j,1}, W_{j,2})$ denote the waiting-time KPIs for the two contact groups at epoch $j$, 
$\mathbf{A}_j = (A_{j,1}, A_{j,2})$ the corresponding abandonment rates, 
$\mathbf{U}_j$ the vector of expert utilizations, 
and $\mathbf{B}_j$ the vector of remaining back-office tasks.
The reward at epoch $j$ is defined as
\[
R_j
=
-10 \sum_{k=1}^2 \mathbf{I}\{W_{j,k} > 2\}
-40 \sum_{k=1}^2 \mathbf{I}\{W_{j,k} > 4\}
-20 \sum_{k=1}^2 \mathbf{I}\{A_{j,k} > 0.3\}
-60 \sum_{k=1}^2 \mathbf{I}\{A_{j,k} > 0.5\}
-4 \sum_{\ell} \mathbf{I}\{U_{j,\ell} > 0.9\},
\]
where $\mathbf{I}\{\cdot\}$ denotes the indicator function. At the terminal epoch $T$, an additional terminal penalty of $- 20 \sum_{\ell} B_{T,\ell}$ is applied. 

\begin{figure}[htbp]
    \centering

    % --- First row ---
    \begin{subfigure}[t]{0.48\linewidth}
        \centering
        \includegraphics[width=\linewidth]{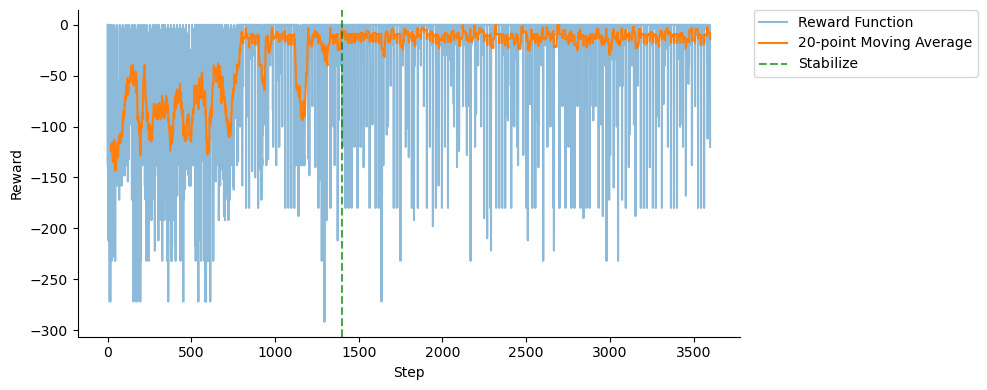}
        \caption{\textbf{Direct simulation-based training.}}
        \label{fig:data}
    \end{subfigure}
    \hfill
    \begin{subfigure}[t]{0.48\linewidth}
        \centering
        \includegraphics[width=\linewidth]{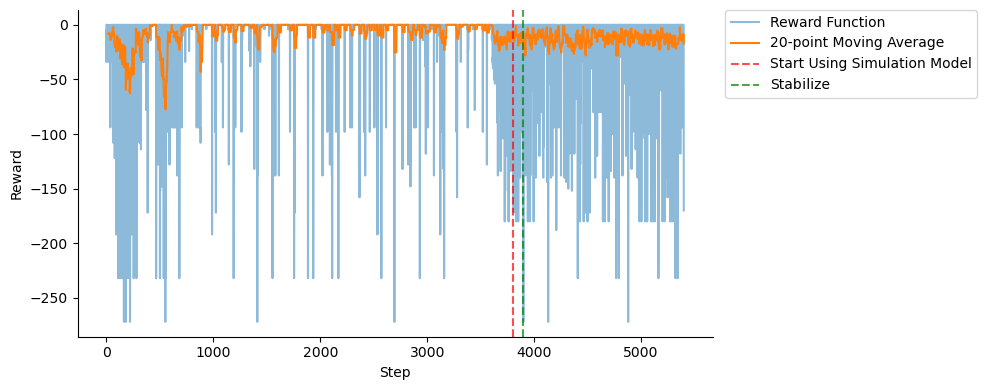}
        \caption{\textbf{Surrogate-based pretraining + simulation fine tuning.}}
        \label{fig:data_surrogate}
    \end{subfigure}

    \vspace{0.5em}

    % --- Second row ---
    \begin{subfigure}[t]{0.48\linewidth}
        \centering
        \includegraphics[width=\linewidth]{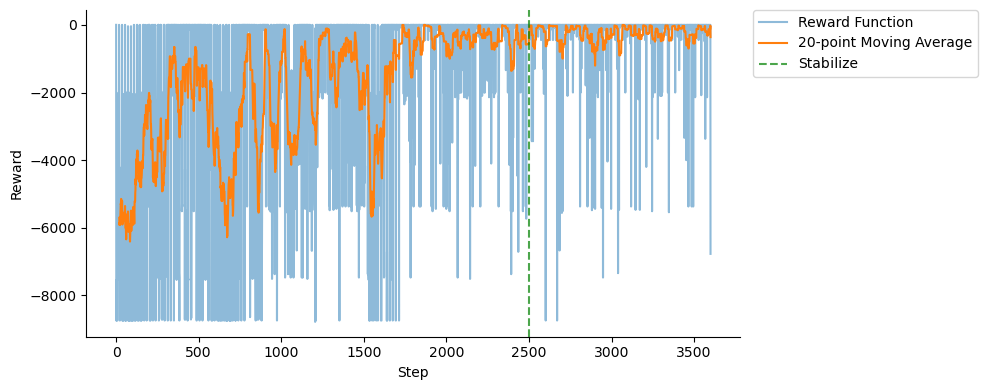}
        \caption{\textbf{Direct simulation-based training after reward change.}}
        \label{fig:data_reward}
    \end{subfigure}
    \hfill
    \begin{subfigure}[t]{0.48\linewidth}
        \centering
        \includegraphics[width=\linewidth]{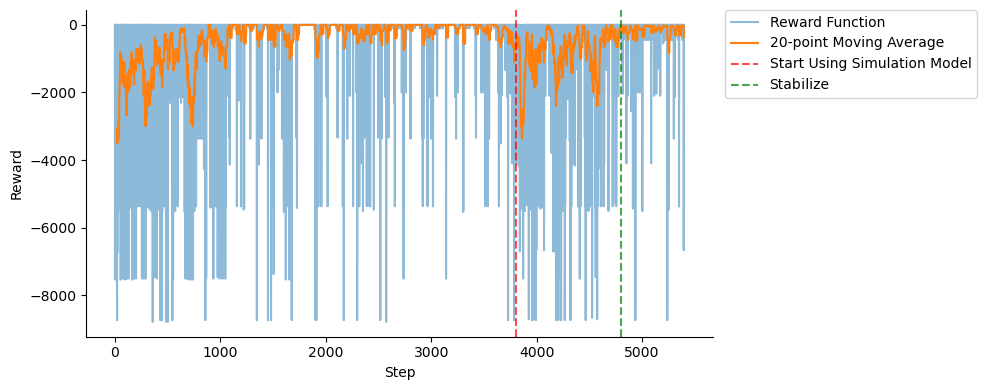}
        \caption{\textbf{Surrogate-based initialization after reward change.}}
        \label{fig:data_surrogate_reward}
    \end{subfigure}

    \caption{
    \textbf{Reward trajectories under original and modified reward structures.}
    Top row: original reward formulation.
    Bottom row: modified reward formulation.
    Left: direct simulation-based training.
    Right: surrogate-based pretraining followed by simulation-based fine tuning.
    }
    \label{fig:reward_comparison}
\end{figure}

A surrogate model is then trained on these sample paths from 200 simulation replications to approximate the simulation response per-epoch in the regions of the state-action space visited by the RL policy. The surrogate model is a neural network consisting of two fully connected hidden layers of width $64$, each followed by a ReLU activation and dropout with rate $0.1$ to mitigate overfitting. To make the model generative, we sample inputs from the fitted input models and incorporate them into the input space, which includes the number of arrivals in the subsequent intervals.
It took 27 seconds to run 200 training epochs for the neural network with a training dataset of 200 simulation replications. Since there are near-zero values in almost all metrics, we choose RMSE as our measure of accuracy. The RMSE for all metrics was below 0.2, except for waiting times, which were about 1 for both contact groups. A more accurate model could be built if either there were more data available or if one builds the surrogate model for stand-alone decision making purposes, where one can choose the thresholds for accuracy. 
Next, a new RL model is trained from scratch using the surrogate model as its environment and is subsequently fine-tuned using the original simulation model. Figures~\ref{fig:data} and~\ref{fig:data_surrogate} compare the learning behavior of the two training strategies.
Surrogate-based initialization leads to substantially faster reward improvement during the early stages of training.
When the RL model is trained using only the simulation model, the reward trajectory requires approximately $80$ simulation replications ($116$ seconds) to stabilize.
In contrast, surrogate-based pretraining using $200$ surrogate replications requires only $10$ seconds, after which fewer than $5$ additional simulation replications are sufficient for the reward to stabilize. This corresponds to approximately a 4-fold reduction in computational cost when accounting for the surrogate model training time, and up to a 10-fold reduction when it is excluded.

We then modify the reward function of the simulation model to reflect a different operational objective.
The reward at epoch $j$ is re-defined as
\[
\begin{aligned}
R_j
=&\;
- \sum_{k=1}^{2} \Big(
100\,\mathbf{I}\{ W_{j,k} > \tau^{(1)}_{W,k} \}
+1200\,\mathbf{I}\{ W_{j,k} > \tau^{(2)}_{W,k} \} \\
&\qquad\qquad
+40\,\mathbf{I}\{ A_{j,k} > \tau^{(1)}_{A,k} \}
+30\,\mathbf{I}\{ A_{j,k} > \tau^{(2)}_{A,k} \} \Big)
-2000 \sum_{\ell=1}^{3} \mathbf{I}\{ U_{j,\ell} > \tau_{U,\ell} \}.
\end{aligned}
\]
The threshold vectors are given by $\boldsymbol{\tau}^{(1)}_W = (1,2)$, $\boldsymbol{\tau}^{(2)}_W = (2,4)$, $\boldsymbol{\tau}^{(1)}_A = (0.2,0.3)$, $\boldsymbol{\tau}^{(2)}_A = (0.5,0.5)$, and $\tau_U = (0.9,0.9,0.9)$.
At the terminal epoch, $T$, an additional penalty of $- 50 \sum_{\ell} B_{T,\ell}$ is applied.
Under this modified reward structure, we retrain the RL model using two strategies:
(i) direct training using the simulation model, and
(ii) surrogate-based pretraining followed by simulation-based fine tuning.
Figures~\ref{fig:data_reward} and~\ref{fig:data_surrogate_reward} show that when the RL model is trained solely using the simulation model, the reward trajectory requires approximately $130$ simulation replications (corresponding to $208$ seconds of wall-clock time) to stabilize.
In contrast, surrogate-based pretraining using $200$ surrogate replications requires only $10$ seconds, after which $55$ additional simulation replications are sufficient for the reward to stabilize, which adds up to $90$ seconds in total, resulting in an approximate two-fold reduction in total training time.

\section{Conclusion}
\label{sec:conclusion}
This paper investigates how simulation surrogate models can accelerate RL training in stochastic systems. We show that when a high-fidelity simulation model is available, surrogate-based pretraining can substantially reduce computational cost while preserving decision quality, as demonstrated through numerical experiments on a stochastic call-center system under both original and modified reward structures.
Several directions for future research are promising.
First, adaptive schemes that dynamically switch between surrogate and simulation environments during RL training merit investigation.
Second, applying the proposed methodology to large-scale systems with high-dimensional state and action spaces, as well as to multi-agent reinforcement learning settings, represents an important avenue for extending the practical impact of surrogate-accelerated RL.
Overall, this work highlights the potential of simulation surrogate models not merely as tools for performance approximation, but as interactive components in accelerating learning-based decision making for complex stochastic systems.

\section*{Acknowledgments}
\label{sec:ack}

This work was supported by the Intuit University Collaboration Program. We also thank Dusan Bosnjakovic for his support throughout the project and Soham Das for helpful conversations.

\bibliographystyle{plain} 
\bibliography{refs}

\end{document}